\documentclass{CUP-JNL-EDS}

\usepackage{latexsym}
\usepackage{graphicx}
\usepackage{multicol,multirow}
\usepackage{amsmath,amssymb,amsfonts}
\usepackage{mathrsfs}
\usepackage{amsthm}
\usepackage{rotating}
\usepackage{appendix}
\usepackage[authoryear]{natbib}
\usepackage{ifpdf}
\usepackage[T1]{fontenc}
\usepackage{times}
\usepackage{newtxmath}
\usepackage{textcomp}%
\usepackage{xcolor}%
\usepackage{hyperref}
\usepackage{lipsum}
\usepackage{float}
\usepackage{subcaption}
\usepackage[justification=centering]{caption}

\jname{Environmental Data Science}
\jyear{2023}
\jvol{4}
\jdoi{10.1017/eds.2023.xx}

\begin{document}

\begin{Frontmatter}

\title[Article Title]{Climate Model Driven Seasonal Forecasting Approach with Deep Learning\footnote{This article is under consideration for 12th International Conference on Climate Informatics 2023.\vspace*{1cm}}
} 

\author*[1]{Alper Unal}\orcid{0000-0002-5870-5789} \email{alper.unal@itu.edu.tr}
\author[2]{Busra Asan}\orcid{0000-0003-1656-986X} \email{asan18@itu.edu.tr} 
\author[1]{Ismail Sezen} \orcid{0000-0003-2260-1846}\email{sezenis@itu.edu.tr}
\author[2]{Bugra Yesilkaynak} \orcid{0000-0003-1649-9865} \email{bugrayesilkaynak@gmail.com}
\author[1]{Yusuf Aydin} \orcid{0000-0002-2310-6778} \email{yusufaydin@itu.edu.tr}
\author[1]{Mehmet Ilicak}\orcid{0000-0002-4777-8835} \email{milicak@itu.edu.tr}
\author[3]{Gozde Unal}\orcid{0000-0001-5942-8966} \email{gozde.unal@itu.edu.tr}

\address[1]{\orgdiv{Eurasia Institute of Earth Sciences}, \orgname{Istanbul Technical University}, \orgaddress{\city{Istanbul},  \country{Turkiye}}}

\address*[2]{\orgdiv{Department of Computer Engineering}, \orgname{Istanbul Technical University}, \orgaddress{\city{Istanbul},  \country{Turkiye}}}

\address*[3]{\orgdiv{Department of AI and Data Engineering}, \orgname{Istanbul Technical University}, \orgaddress{\city{Istanbul},  \country{Turkiye}}}

\received{}
\revised{}
\accepted{}

\authormark{Unal, A. et al.}


\abstract{Understanding seasonal climatic conditions is critical for better management of resources such as water, energy and agriculture. Recently, there has been a great interest in utilizing the power of artificial intelligence methods in climate studies. This paper presents a cutting-edge deep learning model (UNet++) trained by state-of-the-art global CMIP6 models to forecast global temperatures a month ahead using the ERA5 reanalysis dataset. ERA5 dataset was also used for finetuning as well performance analysis in the validation dataset. Three different setups (CMIP6 ; CMIP6 + elevation; CMIP6 + elevation + ERA5 finetuning) were used with both UNet and UNet++ algorithms resulting in six different models. For each model 14 different sequential and non-sequential temporal settings were used. The Mean Absolute Error (MAE) analysis revealed that UNet++ with CMIP6 with elevation and ERA5 finetuning model with “Year 3 Month 2” temporal case provided the best outcome with an MAE of 0.7. Regression analysis over the validation dataset between the ERA5 data values and the corresponding AI model predictions revealed slope and $R^2$ values close to 1 suggesting a very good agreement. The AI model predicts significantly better than the mean CMIP6 ensemble between 2016 and 2021. Both models predict the summer months more accurately than the winter months.}

\policy{This paper discusses the use of novel developments in machine learning field in seasonal forecasting. Traditionally, climate models are used to simulate physical, chemical, and biological processes in the atmosphere to generate climate projections. Machine learning methods are gaining popularity among different fields. A team of computer scientists and earth scientists worked on this paper that investigates the use of machine learning algorithms along with climate models for seasonal forecasts, which are critical for better resource management.}

\end{Frontmatter}

\section{Introduction}

Seasonal forecast is defined as a variety of potential climate changes that are likely to occur in the coming months and seasons \cite{Pan2022}. This is crucial for governments and decision makers to better manage natural resources such as water, energy, agriculture, as well as protect human health \cite{Yuan2019}. For example, crop producers use seasonal forecasts to make decisions about the timing of planting and harvesting, field fertilization, and water management \cite{Klem2017}.  Weather plays an important role in energy supply and demand \cite{Felice2015}, hence an accurate forecast of the future weather conditions could increase the effectiveness and dependability of energy management at the local and national levels given the requirement to maintain the balance between electricity production and demand. Accurate forecasting of extreme events such as storms, heatwaves, droughts, and floods is required to improve disaster preparedness \cite{Liu2022}.

Weather prediction at shorter time-scales such as daily to monthly depends on understanding of physical processes in the atmosphere as well as interactions between the atmosphere, oceans and land. Scientifically, forecasting for longer time scales is based on the same laws of   physics   as   forecasting   for   shorter   time   scales  \cite{Klem2017}.   An important distinction must be made between dynamical predictions, which use intricate physical numerical models,   and   statistical predictions,   that   use   regional   historical   relationships   between   physical   variables   like temperature and precipitation \cite{Franzke2022}, \cite{Klem2017}, \cite{Troccoli2010}, \cite{Roads2003}. 

Two approaches for constraining climate predictions based on past climate change includes large ensembles of simulations from computationally efficient models and small ensembles of
simulations from state-of-the-art coupled ocean-atmosphere General Circulation Models (GCMs) \cite{Stott2007}. GCMs are frequently used in studies related to the
impacts of large-scale climate change \cite{Fujihara2008}. High-resolution climate data from current global climate models are provided using
Regional Climate Downscaling (RCD) techniques \cite{Scinocca2016}, \cite{Laprise2008}. Several programs such as THORPEX, DEMETER, and EUPORIAS have been launched in practice to work toward seasonal forecasting \cite{Klem2017}, \cite{Toth2007}.

 In 1995, Coupled Model Intercomparison Projects (CMIP) began as a comparison of a few pioneering global coupled climate models and outputs are used by different organizations around the world such as IPCC to better understand past, present, and future climate change \cite{Xu2021}, \cite{Wang2021}, \cite{Liu2022}. CMIP6 is the most recent phase of the CMIP. The CMIP6 platform began in 2015, offers the most up-to-date multi-model datasets. Simulation outputs from more than 100 different climate models produced by more than 50 different modeling groups contributed to CMIP6. In addition to historical studies, seasonal forecast for different emission scenarios are provided \cite{Liu2022}, \cite{Zhang2021}, \cite{Turnock2020}, \cite{Fan2020}.

In recent years, big data, effective supercomputers with Graphics Processing Units (GPUs) and scientific interest in novel algorithms and optimization techniques proved to be significant turning points in machine learning history. Machine Learning has recently been a hot topic in climate studies. \cite{Tyagi2022} reviewed a number of studies that applied the different Machine Learning/Deep Learning algorithms in Flash Drought (FD) studies. \cite{Luo2022} used a Bayesian Deep Learning approach to near-term climate prediction in the North Atlantic. \cite{Bochenek2022} investigated the top 500 scientific articles about machine learning in the field of climate and numerical weather prediction that have been published since 2018. \cite{Anochi2021} evaluated different Machine Learning methods for precipitation prediction in South America. \cite{Zhang2021} and \cite{Feng2022} used Deep Learning algorithms to down scale hydro-climatic data of CMIP6 simulations in China. 

This study aims to improve seasonal temperature forecasts using both atmospheric models and machine learning algorithms. Specifically, the objective is to utilize the power of CMIP6 physical models along with European Centre for Medium-Range Weather Forecasts (ECMWF) Reanalysis 5th Generation (ERA5) dataset (created using data assimilation and model forecasts) while utilizing specifically deep neural network learning methods for better global seasonal forecast of 2m temperature. 
 \vspace{-0.2cm}
\section{Materials and Methods}
\subsection{Training data}

This study has utilized the 2m temperature outputs from 9 fully coupled Earth System Models (ESM) that participated in the CMIP6 (Eyring et al. 2016). These models are ACCESS-CM2, CNRM-CM6-1-HR, GISS-E2-1-H, NorESM2-MM, CESM2-WACCM, EC-Earth3-Veg, MPI-ESM1-2-HR, MIROC-ES2L, and IPSL-CM6A-LR.  For each ESM model, only one ensemble member was used and the time slice for the coupled models is chosen from the historical period (i.e., from 1850 to 2014). 

For validation and fine-tuning, 2m temperature data from the ERA5 reanalysis dataset \cite{bib3} have been used. The ERA5 dataset is partitioned into two different time slices: 1973 to 2016 are used for fine-tuning, while 2016 to 2021 are used for evaluation of the trained models. In order to estimate the performance of the deep learning model (which is called as AI model from now on) between 2016 and 2021, we also used the multi-model mean of the CMIP6 models using the IPCC SSP5-8.5 scenario which is selected since there was no significant reduction in carbon emissions after 2014 when the historical simulations ended.
 
\subsection{Model Architecture}
\label{sec:ModelArchitecture}
As this study focuses on spatio-temporally-varying data, an encoder-decoder based architecture UNet++ \cite{Zhou2018}, which is based on the original UNet architecture \cite{Ronneberger2015} is adapted. In UNet++, the skip connections of the UNet are re-designed to minimize the semantic gap between the feature maps coming from the sub-networks of the encoder and the decoder, making the learning easier. Our model specifically employs Convolutional Neural Network (CNN) layers due to nature of the input data.  We construct the UNet++ in order to perform a prediction task, which is explained below. The architecture includes a contracting path, i.e. an encoder part, which summarizes the information by reducing the size of the input image and increasing the number of channels. This downsampling operation results in spatial information loss due to the compression of the input. The UNet++, as in the original UNet, introduces skip connections that reduce the information loss after the bottleneck layer and recovers fine-grained details. Skip connections aggregate information from different resolution levels in order to increase accuracy and speed up the convergence. In the expansive path, i.e. the decoder part, skip connections concatenate the outputs of each downsampling layer to corresponding upsampling layers, aiming at image reconstruction that is at the same spatial resolution as that of the input. 

\begin{figure}[t]
\centering

\begin{subfigure}[b]{1\columnwidth}
  \includegraphics[width=\linewidth]{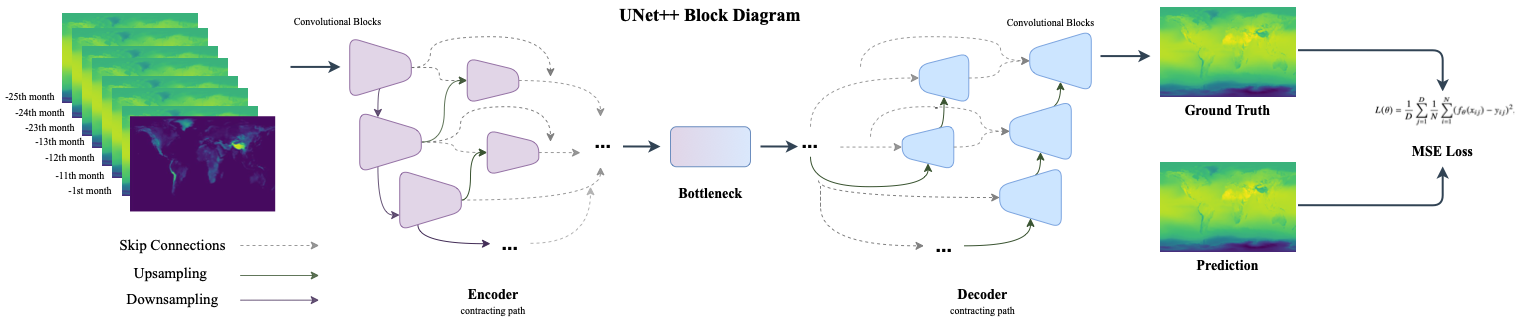}
 \caption{Network architecture of the proposed model.} 
 \label{fig:architecture}
\end{subfigure}

\begin{subfigure}[b]{1\columnwidth}
  \includegraphics[width=\linewidth]{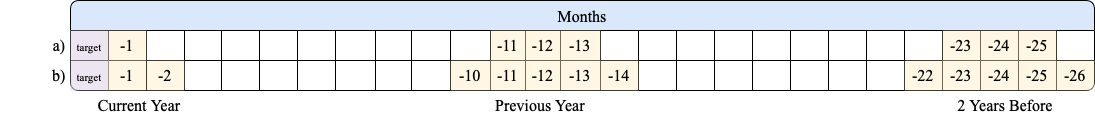}
  \caption{A sample arrangement of the months for the second experimental setting.}
  \label{fig:exp_setting}
\end{subfigure}

\caption{$a)$ Depiction of the UNet++ model which we adapted to our seasonal temperature forecast task. Month descriptions in the input of the encoder refer to relative timing of the input channels (e.g. each month used) according to the target month. In addition to input months, an elevation map is added as a separate channel. $b)$ Arrangement of the months for the multi-dimensional input for the experimental settings: $a)$ 2 years 1 months (given in the first row), and $b)$ 2 years 2 months (given in the second row) are shown}
\end{figure}

Figure~\ref{fig:architecture} depicts the block diagram of the neural network model that we construct for the seasonal forecast of temperature. In addition to 2m temperature, an elevation map is stacked to the input as ancillary data to investigate whether it can improve the prediction performance. 
We use two different experimental settings to evaluate the effectiveness of using fully historical input data versus periodic data. The first setting is designed to see the performance of the historical data stacked sequentially in a temporal manner such as from $t-1$ to $t-6$, $t-12$ and up to $t-36$. Whereas in the second set, the target month and its neighbours are stacked in a yearly manner from historical data to assess the effect of periodicity. Months before the target prediction, the previous years' target month and months before and after them (e. g. neighboring months) are stacked as the input. The number of previous lag years as well as the number of preceding and succeeding  months are selected as hyper-parameters. We have explored lag years 1 to 4 and preceding/succeeding 1 and 2 months. Two examples for these settings are depicted in \ref{fig:exp_setting}. The input to the Encoder network consists of the maps (temperature and elevation) corresponding to the month $t-1$, which is one previous to the target month $t$, $t-12 ~\pm \Delta t$, $t-24 ~\pm \Delta t$, and $t-36 ~\pm \Delta t$, where $\Delta t$ could be either one of $\{1,2\}$. Specifically, we selected $\Delta t = 2$ in order to account for possible seasonal monthly shifts. The overall concatenated input tensor goes through the UNet++ model, and a single prediction map for temperature at the target month $t$ is produced at the output of the network, as visualized in Figure~\ref{fig:architecture}.
Investigating the UNet and its successor the UNet++ with ERA5 fine-tuning and including elevation information resulted in 6 different AI Models to train. Considering all hyper-parameters (i.e., sequential and non-sequential), we have designed 14 experimental settings (named as cases from now on) resulting in 84 simulations in total. It should be noted that we also explored going back to lag years from 5 to 10, however as the number of lag years are increased, the amount of validation data is naturally decreased, and as a result, the data size was not adequate for the model optimization process. Therefore, we have stopped at year 4 for model setup.

As the spherical earth in 3D (3-Dimensions) is represented over a 2D spatial grid, one has to pay attention to the the spatial information at the edges of images while applying convolutions. Rather than traditional 3x3 spatial convolutions, 3x3 circular convolutions which pad the input with information from the opposite sides of the image are used to preserve the spatial information at the edges of the image. During downsampling, 3 maxpool operations and 8 convolutional layers with batch normalization are used. Similarly, the upsampling path is designed using 3 upsample and 7 convolutional layers with batch normalization. As UNet++ introduces intermediate feature maps for the skip connections in each level during downsampling and upsampling, 6 convolutional layers are used for constructing all the intermediate feature maps. Moreover, concatenating lower resolution feature maps requires the usage of 3 additional upsample layers. For training the neural network model, as the loss function, the Mean Squared Error (MSE) Loss in (\ref{MSE}) is utilized:
 \vspace{-0.1cm}
\begin{equation}
    L(\theta) = \frac{1}{N}\sum_{i=1}^{N}(f_{\theta}(X_{i}) - Y_{i})^2,
\label{MSE}
\end{equation}
 where $f_{\theta}$ represents the Neural Network model, $X_{i}$ is the input multi-channel tensor consisting of stacked monthly data and elevation data,  $Y_{i}$ is the target temperature in the grid ("Ground Truth"). N corresponds to the number of target time steps in a given batch of the selected stochastic gradient descent optimizer.


\begin{NoHyper}
 \begin{table}[t]
\tabcolsep=0pt%
\TBL{\caption{Mean Absolute Error (MAE) values as estimated for the entire domain (lat:192 x lon:288) for each simulation conducted: 6 models x 14 cases = 84 simulations}}
{\begin{fntable}
\begin{tabular*}{\textwidth}{@{\extracolsep{\fill}}lcccccc@{}}\toprule%
 & \multicolumn{3}{@{}c@{}}{\TCH{CMIP6 Train}}& \multicolumn{3}{@{}c@{}}{\TCH{CMIP6 Train + ERA5 finetune}}
 \\\cmidrule{2-4}\cmidrule{5-7}%
\TCH{Temporal Setting} & \TCH{M1} & \TCH{M2} & \TCH{M3} &
\TCH{M4} & \TCH{M5} & \TCH{M6} \\\midrule
{\TCH{6 months}}&1.136 &1.123 &1.162 &1.030 &1.077 & 1.046\\
{\TCH{12 months}}&1.031 &1.024 &1.014 &1.004 &1.028 & 1.002\\
{\TCH{18 months}}&1.020 &1.027 &1.013 &0.993 &0.997 & 0.999\\
{\TCH{24 months}}&0.998 &0.992 &0.984 &0.981 &0.987 & 0.983\\
{\TCH{30 months}}&1.007 &1.006 &0.989 &1.008 &1.002 & 0.976\\
{\TCH{36 months}}&0.991 &0.979 &0.984 &0.990 &0.981 & 1.003\\
{\TCH{1 year 1 month}}&1.024 &1.024 &0.989 &0.999 &0.989 & 0.983\\
\TCH{1 year 2 months}&1.011 &1.008 &1.008 &0.997 &0.986 & 0.984\\
{\TCH{2 years 1 month}}&0.999 &0.995 &0.996 &0.977 &0.967 & 0.967\\
\TCH{2 years 2 months}&0.996 &0.986 &0.980 &0.973 &0.969 & 0.979\\
{\TCH{3 years 1 month}}&0.981 &0.977 &0.986 &0.988 &0.973 & 0.967\\
\TCH{3 years 2 months}&0.995 &0.989 &0.978 &0.993 &0.969 & 0.975\\
{\TCH{4 years 1 month}}&0.982 &0.989 &0.966 &0.982 &0.965 & 0.962\\
{\TCH{4 years 2 months}}&0.998 &0.984 &0.967 &0.992 &0.965 & 0.955\\
\botrule
\end{tabular*}%
\footnotetext[2]{{Note:} M1 and M4 are UNet; M2 and M5 are UNet with elevation; M3 and M6 are UNet++ with elevation.}
\end{fntable}}
\label{table:experimentsMAE}
\end{table}
\end{NoHyper}
 
\subsection{Training and Evaluation}
\label{train-eval}
After the neural network model is constructed, it is trained with the MSE loss function in all the experiments. A learning rate of $1e-5$ and a weight decay of $1e-3$ is used with a step learning rate scheduler with Adam optimizer \cite{Kingma2015}. The model is optimized for 40 epochs and early stopping is used to avoid over-fitting.  After each convolutional layer, a batch normalization layer is used. Batch size is chosen as 16. Training process is done on an NVIDIA RTX A5000 GPU, and the results are delivered on average after 3-4 hours of training.
\vspace*{-0.05cm}

During validation, the loss versus iterations are monitored, and the model with the lowest validation error is selected. During inference/test time, the input is formed by the monthly temperature data coming from CMIP6 temperature maps that are stacked and given to the model in a feedforward evaluation. In each evaluation experiment, the target is defined as the month after the latest month in the input. After the training with CMIP6 data, the selected model is further fine-tuned with ERA5 t2m data in order to increase the capability of the model in real world forecasting scenarios. Fine-tuning process is carried out by following the same process as in the training. Monthly temperature data taken from ERA5 is stacked as a multi-channel input and given to the network. 

As the evaluation metric, Mean Absolute Error (MAE) (\ref{MAE}) is chosen, where each $x_{ij}$ and $y_{ij}$ corresponds to the predicted temperature and the ground truth temperature value of the corresponding grid, respectively. D refers to the number of longitudes, and M refers to the number of latitudes in the 2D spatial grid. All MAEs are summed and averaged across the temperature map to measure the overall error:
\begin{equation}
     MAE = \frac{1}{D} \sum_{j=1}^{D} \frac{1}{ M}\sum_{i=1}^{M}|x_{ij}-y_{ij}|.
    \label{MAE}
\end{equation}

 \begin{figure}
\centering
\includegraphics[width=0.6\linewidth]{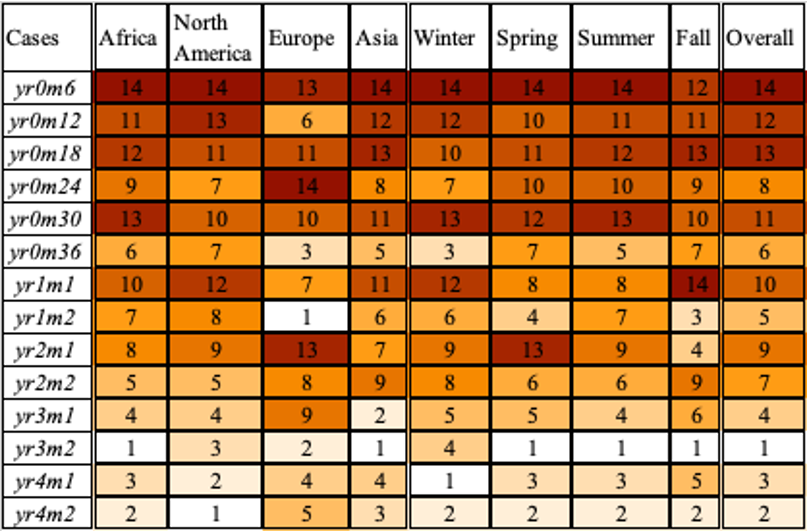}
\caption{MAE Ranks of Model 6 for 14 temporal cases over four continents and four seasons} 
\label{fig:MAERank}
\end{figure}
\vspace*{-0.5cm}

\section{Results and Discussion}
Average MAE results over the validation dataset for temporal cases are provided in Table \ref{table:experimentsMAE}. As seen in this table, MAE ranges between 0.955 and 1.162. For comparison purposes, the MAE of the persistence forecast test (over the ERA5 validation dataset by copying the previous month temperature value as the target month's prediction) is estimated as 2.62. This indicates that all models have improved the MAE significantly with respect to the persistence forecast baseline. It should be noted that out of six  models, for each temporal case, 50 percent (or 7 cases) of the lowest MAEs occur for Model 6 (M6). Three of the minimum MAEs occur for M5, and another three cases occur for M4, whereas only one minimum MAE occurs for  M2. It is clear that ERA5 finetune has improved the performance significantly and using UNet++ with elevation is the best available model. Therefore Model 6 was selected for the rest of the analysis.
 In order to choose the best temporal case for Model 6, we estimate the MAEs and rank them for four main continents (i.e., Africa, North America, Europe, and Asia) as well as their distribution among different seasons (i.e., Winter, Spring, Summer, and Fall). These values along with overall (as estimated over continents and seasons) are given in Figure \ref{fig:MAERank}. As seen in this figure, sequential cases (such as month 6, month 12, even month 36) have higher ranks (hence lower performances) as compared to non-sequential cases. This is possibly due to the fact that the latter is able to recognize the strong seasonality in the data while sequential cases lack this ability. Among the non-sequential cases "Year 3 Month 2" case has the best performance as it has the best MAE rank for Fall, Spring, and Summer (and fourth for Winter). This case is the best for Africa and Asia continents, second for Europe and third for North America. Overall rank for this case is estimated to be number one as well. Therefore, "Year 3 Month 2" is selected as the temporal case for Model 6 (CMIP6 with ERA5 finetune with UNet++ with elevation) for the rest of the analysis.

\begin{figure}[t]
\centering
  \includegraphics[width=0.7\linewidth]{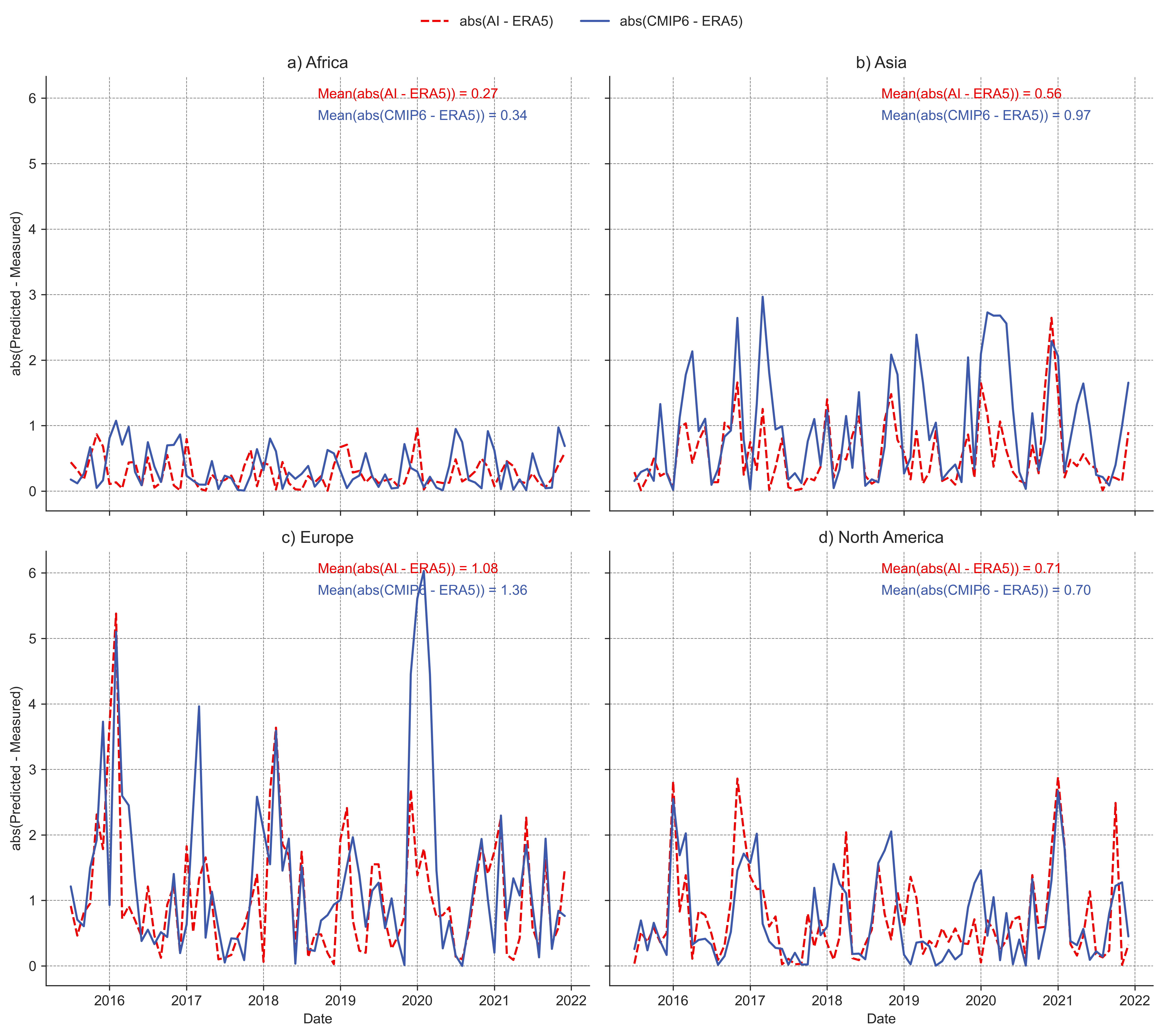}
 \caption{MAE results of AI and CMIP6 models for four different continents as estimated over the validation dataset} 
 \label{fig:errorintime}
\end{figure}

 We compare the performance of the selected AI model to the ensemble mean of the CMIP6 models. Figure \ref{fig:errorintime} shows the MAE for the AI model and the mean of the CMIP6 ensemble as a function of time for the selected continents during the study period from 2016 to 2022. In Africa, both the AI and the CMIP6 model means have a similar error level (i.e., 0.27 versus 0.34) (Fig. \ref{fig:errorintime}-a). This is probably due to the fact that the climate in Africa is fairly uniform as a function of latitude, and both models capture the overall climatology well. The AI model performs better compared to the mean CMIP6 in Asia and Europe. Although the inter-annual variability in the error between two different models overlaps, AI has significantly lower bias values. We believe that since the AI model has been trained by the CMIP6 ensemble, the model might inherit similar inter-annual variability. AI model's MAE value for February 2020 for Europe, Asia, and North America is significantly lower as compared to the CMIP6 model (a difference of 3.2 degrees for Europe, almost 1 degree for Asia, and 0.5 degrees for North America), while they estimate closer values for Africa.  The error in North America is the only place where the CMIP6 ensemble ($0.7$) is slightly better than the AI model ($0.71$) that needs further investigation. In both models, the error increases in the winter months, indicating that the models do not accurately represent the cold climate in the northern hemisphere.

\begin{figure}[ht]
\centering
  \includegraphics[width=1.05\linewidth]{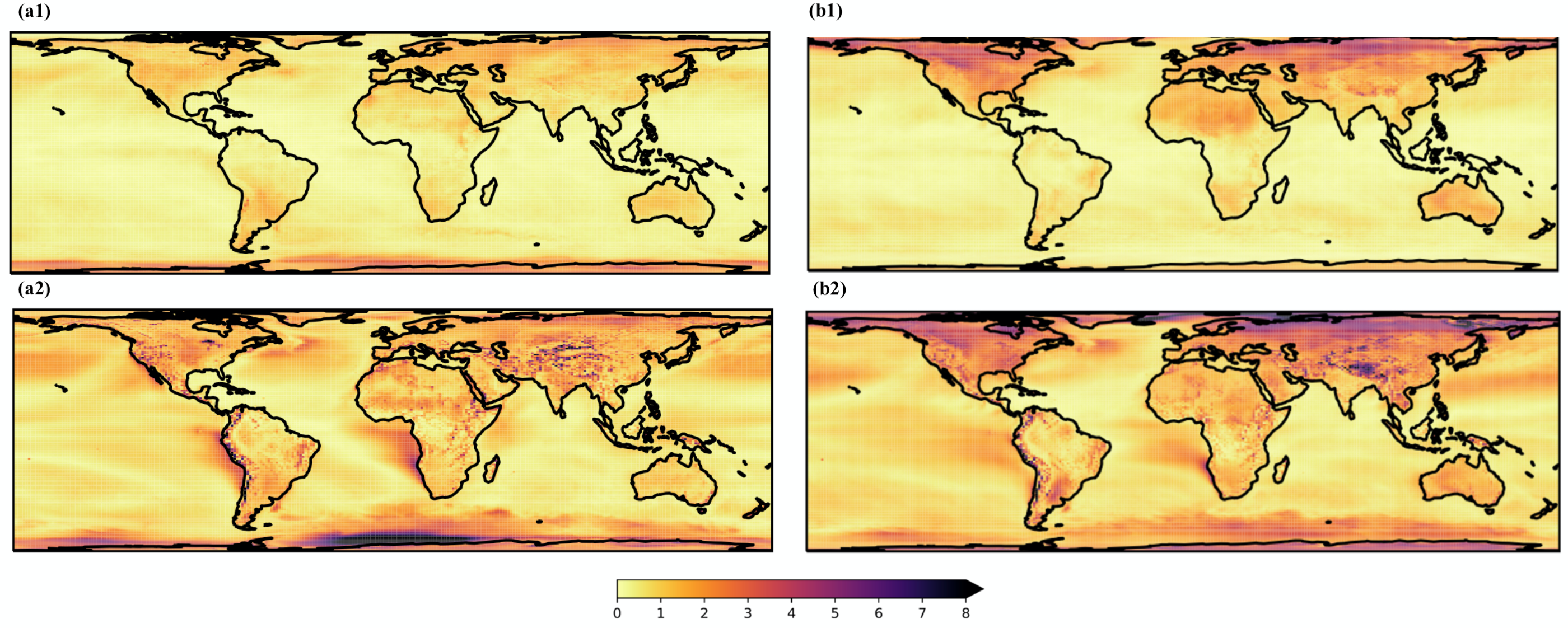}
 \caption{(a) MAE fields of AI model in Summer (a1); CMIP6 model in Summer(a2); AI model in Winter (b1); and CMIP6 model in Winter (b2) for the validation dataset} 
 \label{fig:meanerror}
\end{figure}

The spatial distributions of the MAE fields for summer and winter for the AI and mean CMIP6 models are shown in Figure \ref{fig:meanerror}. Summer time mean absolute error in the AI model is fairly uniform and approximately $1.5^\circ$C over the continents (Fig. \ref{fig:meanerror}-a1). In contrast, the mean CMIP6 shows a relatively larger error (up to 5$^\circ$C) in high-topography regions such as the Himalayas in Asia, the Andes in South America, the Rockies in North America, and the Alps in Europe (Fig.\ref{fig:meanerror}-a2). The MAE pattern in winter of the AI model is similar to the mean CMIP6 in high latitudes in the northern hemisphere (Figs. \ref{fig:meanerror}b1-b2). This indicates that large-scale Jetstream bias from the CMIP6 models is responsible for the AI's poor performance over Siberia and northern America. Once again, the error in winter is larger than in summer in both models, as we have shown in Fig. \ref{fig:errorintime}. The performance of the AI model in terms of MAE is significantly better than that of the mean CMIP6 in both summer and winter.

\begin{figure}[ht]
\centering
  \includegraphics[width=0.85\linewidth]{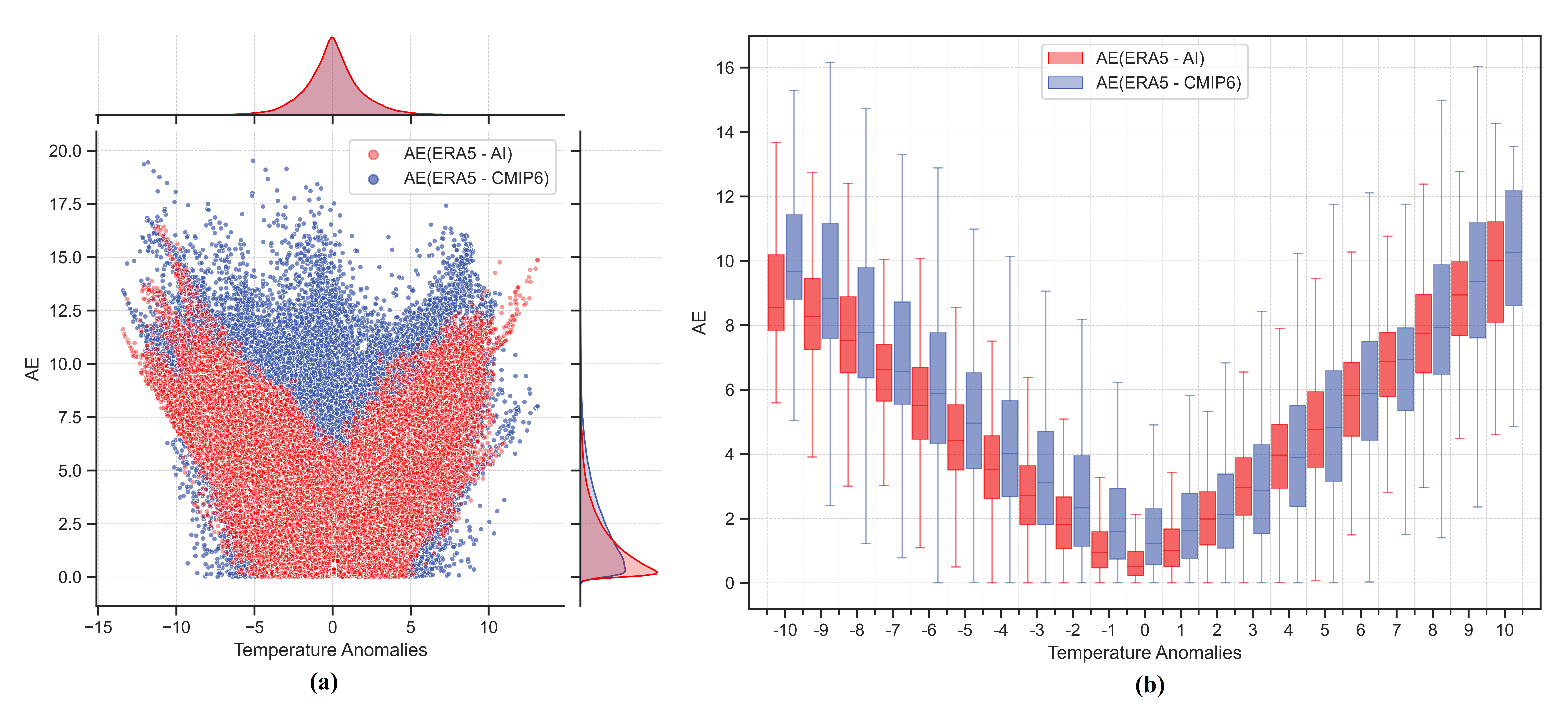}
 \caption{Absolute Error plots of CMIP6 and AI model results for the validation dataset: (a) Scatter (b) Box plots} 
 \label{fig:boxplot}
\end{figure}

Next, at every grid point of the global domain (lat: 192, lon: 288), we calculated the temperature anomalies for each month to remove the mean of the month of that grid point. Then we computed the scatter plot of absolute errors of the mean CMIP6 and AI models for all grids as a function of these temperature anomalies (Figure \ref{fig:boxplot}-a). The AI model performs better when the temperature anomalies are between $-5^\circ$C and $5^\circ$C indicating that if a particular month is around the monthly mean, then the AI model predicts significantly better than the CMIP6 mean. However, if the month is part of an extreme event such as very cold ($\Delta T\approx-10^\circ$C) or very hot ($\Delta T\approx10^\circ$C), AI's performance is getting closer to the CMIP6 mean. To better understand the performance of the selected model, the box plots of the calculated AE (Absolute Error) based on temperature anomalies are given in Figure \ref{fig:boxplot}-b. In all AI versus CMIP6 error bars for each temperature bin, AI model has significantly lower error values (for the median values and $25^{th}$ and $75^{th}$ percentiles). This outcome is even more pronounced especially for the bins between -2 and +2 (as shown on the x-axis).

In addition to scatter plot and box plots of AE values, other statistical values such as $R^2$ are used for understanding the relationship between observed and predicted temperature for all continents based on the results of the selected scenario. The results indicate that predicted values are well-fitting with the observed values in all continents. The $R^2$ values of all continents are close to 1, which imply the power of the AI model in seasonally predicting temperature around the world (see the supplementary figures).

\section{Conclusion}
In summary, we employ an advanced encoder-decoder model (UNet++) trained by state-of-the-art global CMIP6 Earth System models to forecast global temperatures a month ahead using the ERA5 reanalysis dataset. This study is a proof of concept for the use of this model in a complex climate system. We found that the deep learning model predicts significantly better than the mean CMIP6 ensemble between 2016 and 2021. The AI model predicts the summer months more accurately than the winter months, similar to the mean CMIP6. In the future, we plan to develop this new model using additional atmospheric and oceanic variables such as wind velocities, sea surface temperature, and 500 hPa geopotential height to investigate the effects of additional information on the prediction. We also plan to improve our forecast time to seasonal predictions, that is, three months ahead.

\newpage

\paragraph{Author Contributions}
 Conceptualization: A.U; G.U; M.I. Methodology: A.U.; G.U.; M.I.; Software: B.A.; I.S.; B.Y.; A.U.; Data curation: I.S.; B.Y. Data visualisation: B.A.; I.S.; Y.A.; A.U.;  Writing original draft: B.A.; Y.A.; Writing- review and editing: A.U.; M.I.; G.U. All authors approved the final submitted draft.

\newpage

\paragraph{Supplementary Material}
Two figures provided below are intended to be included as the supplementary material.

\begin{figure}[H]
    \centering
     \begin{subfigure}[b]{0.45\textwidth}
         \centering
         \includegraphics[width=1\linewidth]{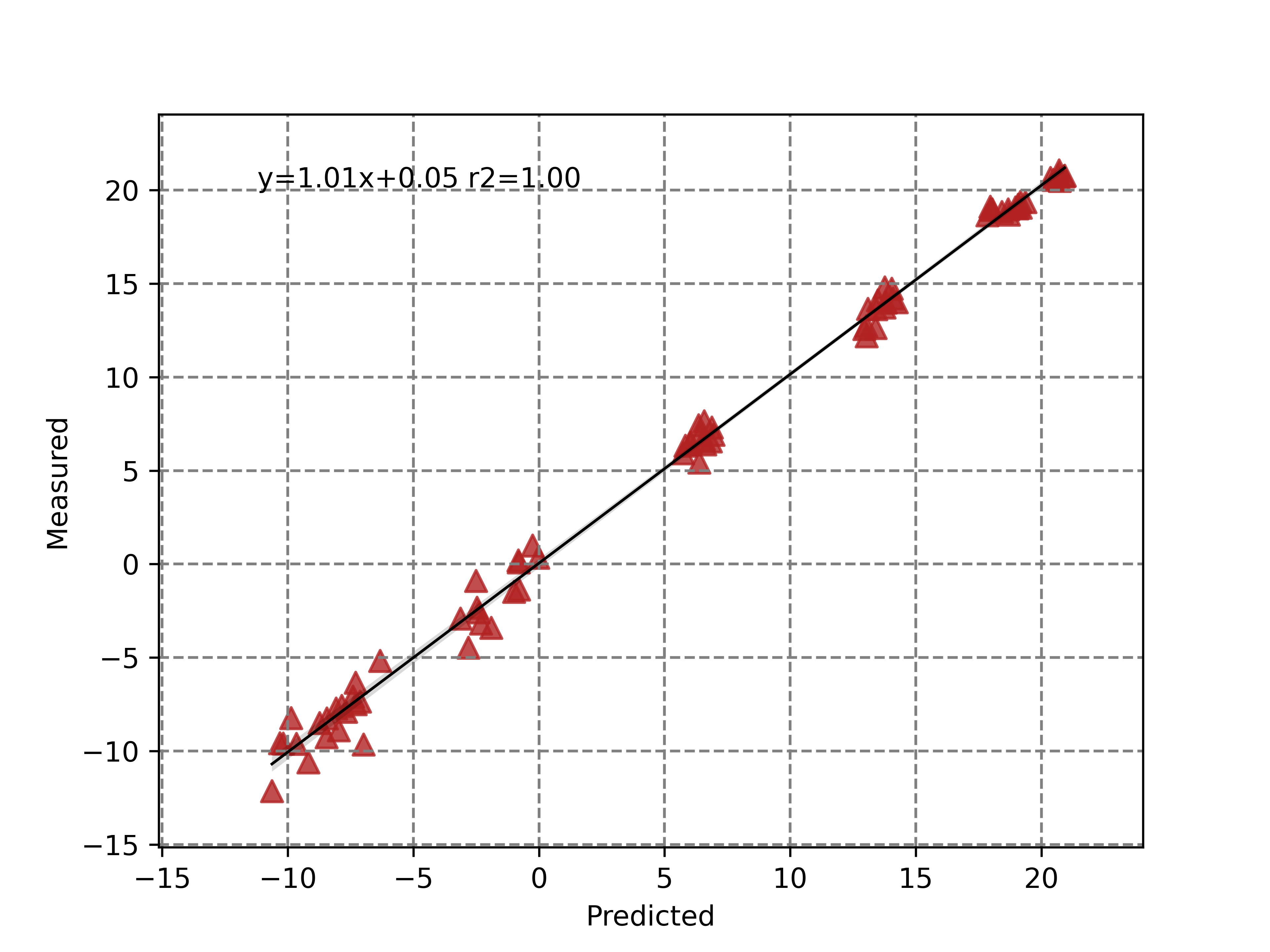}
         \caption{Asia, Scenario 4 year 3 month 2} 
         \label{fig:scatterAsia}
    \end{subfigure}
     \begin{subfigure}[b]{0.45\textwidth}
         \centering
         \includegraphics[width=1\linewidth]{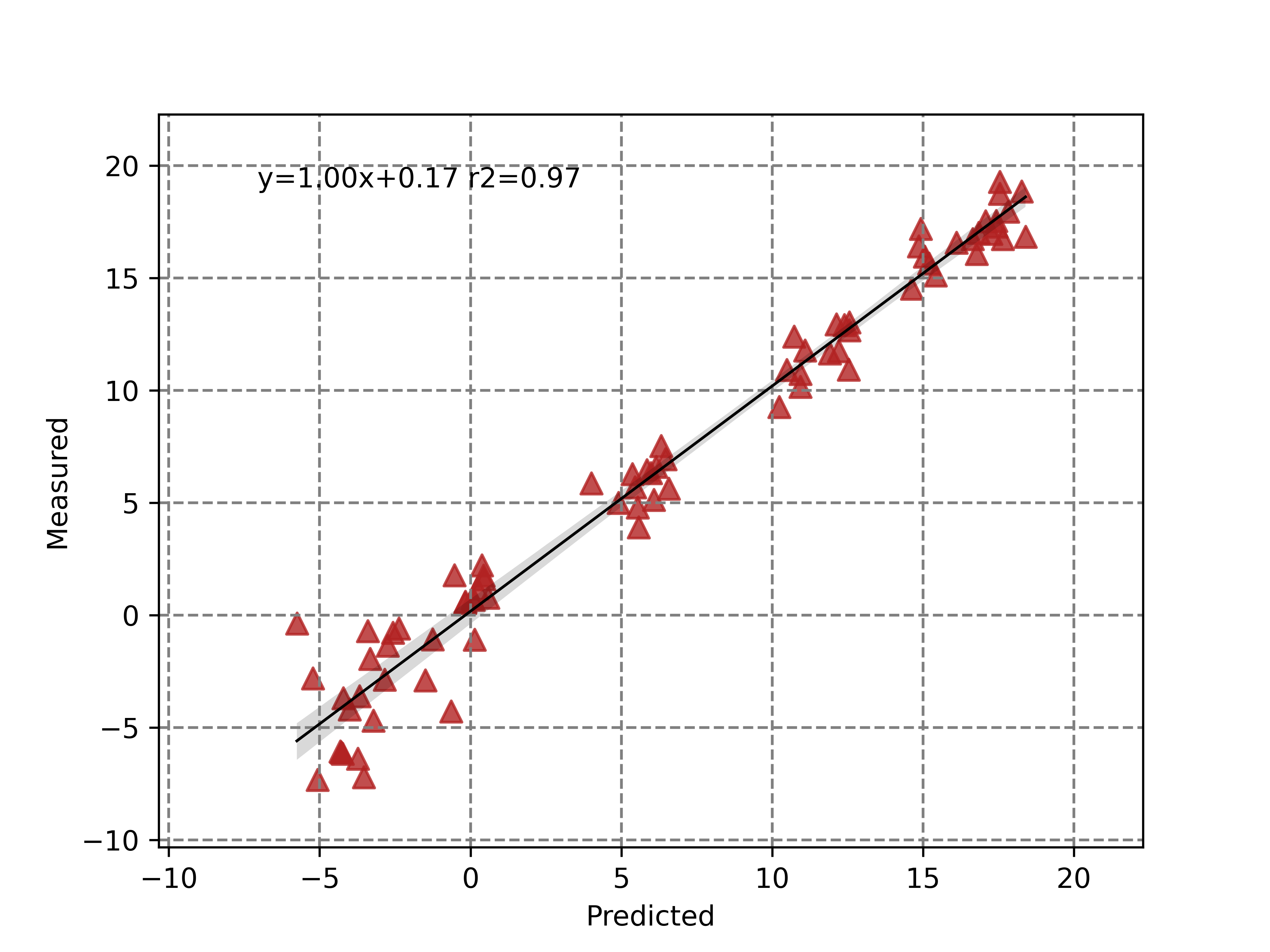}
 \caption{Europe, Scenario 4 year 3 month 2} 
 \label{fig:scatterEurope}
 \end{subfigure}
 \caption{Comparison between AI model and ERA5 for the validation dataset}
\end{figure}


\end{document}